\documentclass[runningheads]{llncs}
\usepackage[T1]{fontenc}
\usepackage{graphicx}
\usepackage{booktabs}
\usepackage[misc]{ifsym}
\newcommand{\corr}{(\Letter)}
\usepackage{float}
\usepackage{longtable}
\usepackage{multirow}
\usepackage{array}
\usepackage{ragged2e}
\usepackage[compatibility=false]{caption} 
\usepackage{mwe}
\usepackage{amsmath}
\usepackage{amssymb}
\usepackage{subfig}
\usepackage{pgfplots}
\pgfplotsset{compat=1.18}
\usepackage{pgfplotstable}
\renewcommand{\figurename}{\textbf{Fig.}}
\usepackage[colorlinks=true, linkcolor=blue, citecolor=blue, urlcolor=blue]{hyperref}

\begin{document}
\justifying
\title{Interpretable Hybrid-Rule Temporal Point Processes}

\titlerunning{Interpretable Hybrid-Rule Temporal Point Processes}

\author{Yunyang Cao \and
Juekai Lin*  \and
Hongye Wang*  \and
Wenhao Li \corr \and
Bo Jin \corr
}

\authorrunning{Y. Cao et al.}

\institute{Tongji University, Shanghai 200092, China 
\email{\{whli, bjin\}@tongji.edu.cn}
}

\tocauthor{Yunyang Cao, Juekai Lin, Hongye Wang, Wenhao Li, Bo Jin}
\toctitle{Interpretable Hybrid-Rule Temporal Point Processes}

\maketitle {\scriptsize* These authors contributed equally to this work.}

\begin{abstract}
Temporal Point Processes (TPPs) are widely used for modeling event sequences in various medical domains, such as disease onset prediction, progression analysis, and clinical decision support. Although TPPs effectively capture temporal dynamics, their lack of interpretability remains a critical challenge. Recent advancements have introduced interpretable TPPs. However, these methods fail to incorporate numerical features, thereby limiting their ability to generate precise predictions. To address this issue, we propose Hybrid-Rule Temporal Point Processes (HRTPP), a novel framework that integrates temporal logic rules with numerical features, improving both interpretability and predictive accuracy in event modeling. HRTPP comprises three key components: basic intensity for intrinsic event likelihood, rule-based intensity for structured temporal dependencies, and numerical feature intensity for dynamic probability modulation. To effectively discover valid rules, we introduce a two-phase rule mining strategy with Bayesian optimization. To evaluate our method, we establish a multi-criteria assessment framework, incorporating rule validity, model fitting, and temporal predictive accuracy. Experimental results on real-world medical datasets demonstrate that HRTPP outperforms state-of-the-art interpretable TPPs in terms of predictive performance and clinical interpretability. In case studies, the rules extracted by HRTPP explain the disease progression, offering valuable contributions to medical diagnosis. The supplementary material is available at https://github.com/yy-c/HRTPP.

\keywords{Temporal point processes \and Interpretability \and Event sequence modeling \and Logic rules}
\end{abstract}

\section{Introduction}
Event sequence data is prevalent in various domains, especially healthcare, including disease onset prediction \cite{che2018recurrent,renc2024zero}, progression analysis \cite{fan2023construction,kuang2024unveiling}, and clinical decision support \cite{hayat2022medfuse,palma2006fuzzy}, where events occur asynchronously over time. Unlike traditional continuous-time series, which record values at regular intervals, event sequence data consists of discrete events with irregular time stamps, highlighting the necessity of specialized methods for discrete temporal data processing. Numerical features often accompany event sequences, providing crucial contextual information such as intensity, magnitude, or categorical attributes, which are essential for capturing complex temporal dependencies and enhancing predictive performance \cite{meng2024transfeat}.

Temporal Point Processes (TPPs) are widely used for modeling event sequence data due to their ability to capture the stochastic nature of event occurrences over time. Traditional point process models, such as the Poisson process \cite{jorgenson1961multiple}, Hawkes process \cite{hawkes1971spectra}, and self-correcting process \cite{isham1979self}, rely on predefined parametric intensity functions, which often struggle to capture complex temporal dependencies and require strong assumptions about the underlying data distribution. To address these limitations, neural point processes, including models based on recurrent neural networks (RNNs) \cite{du2016recurrent}, transformer architectures \cite{zuo2020transformer}, and continuous normalizing flows \cite{yang2019pointflow}, have been proposed to learn flexible, data-driven representations of event dynamics. However, these learning-based approaches often lack interpretability and require large amounts of training data to generalize effectively. 

The interpretability of temporal point processes is crucial for many real-world applications, especially in healthcare, where understanding the underlying event dynamics is as important as making accurate predictions. For example, in medical diagnostics, an interpretable TPP model can assist doctors in analyzing disease progression by identifying key rules influencing the timing of medical events. Recent advancements have introduced interpretable TPPs, incorporating rule-based mechanisms \cite{kuang2024unveiling,li2021explaining,walzer2007temporal} or sparse attention structures \cite{li2023sparse}. However, these methods typically sacrifice key advantages of numerical features, thereby limiting their ability to generate precise predictions. Thus, existing approaches face several limitations: traditional TPP models \cite{di2015compound,hawkes1971spectra}, though capable of utilizing numerical features, often lack the flexibility to capture complex temporal dependencies; neural TPP models \cite{du2016recurrent}, though powerful, are typically considered black boxes with limited transparency; and interpretable models \cite{kuang2024unveiling,li2023sparse}, while offering some level of interpretability, often fail to incorporate rich numerical features effectively, reducing their predictive accuracy and rule reliability.

An urgent issue is ensuring both interpretability and accuracy. To address these issues, we propose Hybrid-Rule Temporal Point Processes (HRTPP), a novel framework that integrates rule-based mechanisms with numerical feature enhancement. HRTPP integrates three key intensity components: basic intensity, rule-based intensity, and numerical feature intensity, which together determine the overall intensity. The numerical feature intensity captures the influence of continuous-valued attributes. The rule-based intensity encodes temporal dependencies through predefined logic rules, which are optimized via a structured rule mining process involving Bayesian optimization. The best rule set is finally selected by a two-stage mining strategy. 

The \textbf{contributions} of this paper are as follows:
\begin{itemize}
    \item[•] A rule-guided numerical augmentation framework is proposed, integrating rule-based mechanisms with numerical features to enhance TPP modeling.
    \item[•] A two-phase rule mining strategy is introduced to balance search space complexity and predictive performance.
    \item[•] A comprehensive evaluation paradigm for interpretable temporal point process models is established, incorporating multiple criteria to assess both predictive performance and rule reliability.
\end{itemize}

\section{Related Work}
\subsubsection{Temporal Point Processes.} Temporal Point Process (TPP) models provide a probabilistic framework for capturing event sequences in continuous time. A key challenge in TPP research lies in designing intensity functions that are both expressive and interpretable. Efforts to improve interpretability in TPPs have led to the development of attention-based and rule-based approaches. Attention-based approaches such as SA-HP \cite{zhang2020self} and THP \cite{zuo2020transformer} incorporate self-attention mechanisms to capture long-range dependencies in event sequences. ITHP \cite{meng2024interpretable} further explains THP by algebraic operations. Although these methods improve flexibility, the trade-off between accuracy and interpretability remains an open problem. Rule-based TPP models focus on enhancing interpretability through structured rule sets. TLPP \cite{li2020temporal} first introduces logic rules to model event intensities and defines temporal rules as a combination of temporal predicates and relations. TELLER \cite{li2021explaining} introduces a principled approach to mine rules that explain temporal dependencies. Based on this, CLUSTER \cite{kuang2024unveiling} and NS-TPP \cite{pmlr-v235-yang24ag} design different deep neural network structures to accelerate rule mining. Clock Logic Neural Network (CLNN) \cite{yan2023weighted} incorporates weighted clock logic formulas as rules to model event interactions. The recent trend in TPP research highlights the need to balance both explainable and effective aspects, integrating structured rule-based approaches with deep learning models for more explainable and effective event modeling.
\subsubsection{Rule Mining.} Rule mining focuses on automatically discovering logical rules from event sequence data to explain event occurrence patterns. Traditional methods for rule discovery primarily identify frequent patterns, but struggle with capturing temporal dependencies. Apriori \cite{agrawal1994} efficiently extracts frequent event sets but fails to model event order. Sequential pattern mining techniques, such as CM-SPADE \cite{fournier2014a} and VGEN \cite{fournier2014b}, attempt to incorporate temporal orders but face difficulties in handling fine-grained timestamps. To overcome these limitations, TELLER \cite{li2021explaining} formulates rule discovery as a maximum likelihood problem to generate high quality rules, but requires a lot of computation. CLUSTER \cite{kuang2024unveiling} introduces an iterative framework based on the Expectation-Maximization algorithm to improve rule mining efficiency. NS-TPP \cite{pmlr-v235-yang24ag} represents predicates and logic rules as vector embeddings and introduces a neural-symbolic framework for rule mining. Clock Logic Neural Network (CLNN) \cite{yan2023weighted} utilizes weighted clock logic formulas to introduce three novel rule forms and incorporate numerical values into the rules. HyperLogic \cite{yang2024hyperlogic} proposes the integration of if-then logic rules within neural network architectures.

\section{Background}
TPPs are widely used to interpret complex event sequences over time. For example, in intensive care unit (ICU) monitoring, physicians analyze the progression of symptoms and lab results to make decisions. Understanding the logic rules among medical events is crucial for accurate diagnosis and treatment. This section briefly reviews temporal point processes and temporal logic rules.

\subsection{Temporal Point Processes}
Temporal Point Process (TPP) is mathematically defined as a stochastic process $\{t_i\}_{i=1}^N$ where each $t_i$ denotes the timestamp of an event, and $N$ is the number of observed events. The core component of TPPs is the conditional intensity function $\lambda(t)$, which defines the instantaneous rate of event occurrence at time $t$, conditioned on the history $ \mathcal{H}_t=\{t_1, t_2, \dots, t_{i}\}$ up to but not including $t$:
\begin{equation}
\lambda(t|\mathcal{H}_t) = \lim_{\Delta t \to 0} \frac{\mathbb{P}\left(N(t+\Delta t) - N(t) = 1 \mid \mathcal{H}_t\right)}{\Delta t}.
\end{equation}
Here, $\Delta t$ is a small time interval, and $N(t)$ denotes the counting process up to $t$. This intensity function governs the likelihood of an event happening at a particular time. The overall dynamics are characterized by the joint distribution. By modeling $ \lambda(t | \mathcal{H}_t) $, TPPs can describe a wide range of event patterns, including those influenced by past occurrences, external features, or domain-specific rules.

\subsection{Temporal Logic Rules}
Temporal logic rules consist of two parts: temporal predicates and relations. Define a set of predicates as $\mathcal{X}=\{X_1, X_2, \dots, X_n, Y\}$, where each predicates $X_u$ is an independent variable, $Y$ is a target variable. Define a set of temporal relations as $\mathcal{C}$, which define how the variables are related over time. Rules are in the form of "if-then". A generalized temporal logic rule $R$ can be expressed as:

\begin{equation}
R: C_i(\psi_1, \psi_2) \to Y,
\end{equation}
where $ C_i \in \mathcal{C} $ represents a temporal relation, and $ \psi_1, \psi_2 $ are expressions that can either be variables from $ \mathcal{X} $ or recursively defined temporal rules. For example, $\psi$ can represent the variable "high heart rate" or the rule "high heart rate before hypertension". The temporal relations between two temporal points can be divided into four types: \textit{and}, \textit{before}, \textit{equal}, \textit{after} \cite{bruce1972model}. Actually, "$X_1$ \textit{before} $X_2$" and "$X_2$ \textit{after} $X_1$" are equivalent, so the final set of temporal relations is $\mathcal{C}=\{and, before, equal\}$.

\section{Methodology}
\begin{figure}[!t]
    \centering
    \includegraphics[width=1.0\linewidth]{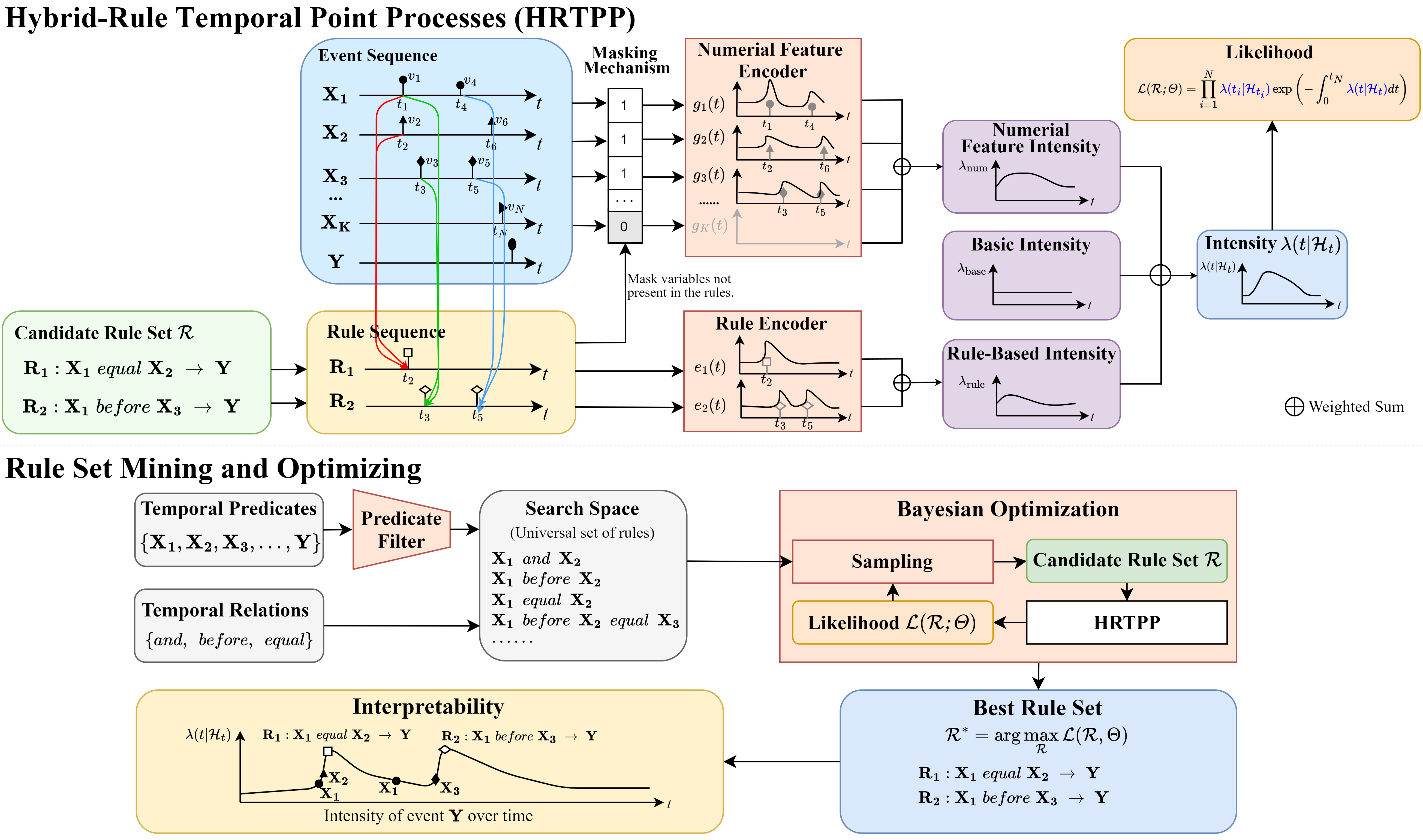}
    \caption{The framework of model. In the HRTPP component, given a candidate rule set and event sequences, the model can compute the likelihood of the intensity function. The model integrates three key intensity components: basic intensity, rule-based intensity, and numerical feature intensity, which together determine the overall event intensity function $\lambda(t | \mathcal{H}_t)$. The numerical feature intensity captures the influence of continuous-valued attributes using a numerical feature encoder and a masking mechanism. The rule-based intensity encodes temporal dependencies through predefined rule set. In the rule set mining and optimizing component, Bayesian optimization iteratively refines the rule set by utilizing HRTPP computed likelihoods to guide the sampling distribution over the search space. The selected rules provide interpretable explanations of critical events via their impact on the intensity function dynamics.}
    \label{fig1}
\end{figure}

This section introduces the framework of Hybrid-Rule Temporal Point Processes (HRTPP), shown in Fig.\@ \ref{fig1}. Given an event sequence $\mathcal{S}=\{(t_i, k_i, v_i)\}_{i=1}^N$ consisting of $N$ events, each event type $k_i \in \{1,2,\dots,K\}$ is associated with an independent variable $X_k$ and a numerical feature $v_i$, providing contextual information. For instance, in ICU monitoring, an event $k_i$ "elevated heart rate" at $t_i = 2$ hours may have $v_i = 120$ bpm. The conditional intensity function $\lambda(t|\mathcal{H}_t)$ is extended to incorporate temporal logic rules alongside numerical attributes, formulated as follows:
\begin{equation}
\lambda(t|\mathcal{H}_t) = f(\mathcal{S}_t; \Theta, \mathcal{R}),
\end{equation}  
where $\mathcal{H}_t$ represents the historical event information, $\mathcal{S}_t=\{(t_i, k_i, v_i)\}_{t_i<t}$ denotes the observed event sequence, $\Theta$ is the set of model parameters, and $\mathcal{R}$ comprises the extracted temporal logic rules that govern event dependencies. 

\subsection{Rule-Based Feature Construction}
Temporal logic rules are essential in medical diagnosis, particularly in ICU monitoring, where capturing symptom progression is crucial for informed clinical decision-making. We propose a rule-based feature construction framework that integrates structured temporal rules with numerical features. This approach encodes event dependencies into numerical representations, ensuring a balance between explainability and data-driven modeling.

\subsubsection{Definitions of Temporal Rules.}
Temporal rules formally define the logical relationships between different events over time, which consists of temporal predicates and temporal relations. In this study, we categorize the temporal relations between two time points into three types: \textit{before}, \textit{equal}, \textit{and}. 

Given two predicates $X_u$ and $X_v$ with timestamps $t_u$ and $t_v$, along with a target event $Y$ occurring at $t_y>\max(t_u,t_v)$ and the indicator function $\mathbb{I}(\cdot)$, the three types of temporal rules are mathematically defined as follows. \textit{Before}, where $X_u$ occurs before $X_v$ within a small time tolerance $\delta$, formulated as $\mathcal{C}_{before}(X_u, X_v) = \mathbb{I}(t_u - t_v < -\delta)$. \textit{Equal}, where $X_u$ and $X_v$ occur approximately at the same time, formulated as $\mathcal{C}_{equal}(X_u, X_v) = \mathbb{I}(|t_u - t_v| \leq \delta)$. \textit{And}, which simply states that both events occur without temporal constraints, formulated as $\mathcal{C}_{and}(X_u, X_v) = 1$. These temporal rules provide a structured mechanism for capturing event dependencies.

\subsubsection{Rule Encoder.}  
To effectively incorporate temporal rules into the TPP framework, we encode hybrid-rule information into a continuous vector space. Based on the definitions of temporal rules, we construct hybrid-rule event sequences. For instance, consider the temporal rule "$X_u$ \textit{before} $X_v$ $\to$ $Y$". The corresponding hybrid-rule event sequence records the timestamps when the condition $X_u$ \textit{before} $X_v$ is satisfied. Specifically, this sequence tracks each moment where the constraint $\mathcal{C}_{before}(X_u, X_v)$ transitions from 0 to 1. The encoding function for a rule $R_j$ is defined as:  
\begin{equation}
    e_{j}(t) = \sum_{t_j \in \mathcal{T}_j} d_\text{rule}(t - t_j),
    \label{eq:ruleemb}
\end{equation}  
where $d_\text{rule}$ is the decay function, and $\mathcal{T}_j$ denotes the set of timestamps when rule $R_j$ is triggered. The choice of the decay function depends on real-world dynamics. For example, an exponential decay function, defined as $d_{rule}(t) = e^{-t}$ for $t \geq 0$, can model the diminishing influence of past rule activations over time. 

\subsubsection{Numerical Feature Encoder.}  
Numerical features provide quantitative insight into the condition of a patient. The numerical feature encoder captures the dynamic influence of numerical attributes over time. Given an event sequence $\mathcal{S}=\{(t_i, k_i, v_i)\}_{i=1}^N$, the sequence corresponding to the predicate $X_k$ is expressed as $\{(t_i, k_i, v_i)\}_{k_i=k}$. Generally, not all predicates directly influence the target event. Therefore, we introduce a masking mechanism to filter out irrelevant predicates. The mask is defined according to the given rule set, where only the predicates appearing in rules associated with the target event remain active, ensuring that the retained variables are indeed the relevant ones. Formally, the mask set is denoted as $\mathcal{M}=\{m_k\}_{k=1}^K$, and the valid predicate set $\mathcal{X}_V \subset \mathcal{X}$ denotes the collection of all variables appearing in the rules. The mask function is defined as $m_k=1$ if $X_k\in\mathcal{X}_V$, and $m_k=0$ otherwise. Using this masking mechanism and the sequence $\{(t_i, k_i, v_i)\}_{k_i=k}$, we define the numerical feature encoding function for predicate $X_k$ as:  
\begin{equation}
    g_{k}(t) = m_k \sum_{i:k_i=k} v_i d_\text{num}(t - t_i),
    \label{eq:numemb}
\end{equation}
where $d_\text{num}$ is a time decay function, that models the influence of past numerical observations. The mask $m_k$ ensures consistency within the HRTPP framework.

\subsection{Hybrid-Rule Temporal Point Processes}
HRTPP extends classical TPP by integrating rule-based dependencies and numerical features. The overall intensity function is composed of three key components. First, basic intensity models fundamental event patterns. Second, rule-based intensity encodes temporal logic rules. Third, numerical feature intensity incorporates clinical measurements to refine event prediction.

\subsubsection{Basic Intensity.}
The basic intensity component $ \lambda_{\text{base}} $ captures the time independent information. This component reflects the inherent likelihood of events occurring. Mathematically, the basic intensity is expressed as: $\lambda_{\text{base}} = \lambda_0$, where $\lambda_0$ is a learnable parameter.

\subsubsection{Rule-Based Intensity.}  
The rule-based intensity component integrates domain expertise and temporal dependencies into the event modeling process. Given a predefined set of temporal rules $\mathcal{R}$, the rule-based intensity function is defined as:  

\begin{equation}  
\lambda_{\text{rule}}(t) = \sum_{R_j \in \mathcal{R}} \alpha_j e_j(t), 
\label{eq:rule}
\end{equation}  
where $\alpha_j$ is a learnable parameter representing the weight associated with rule $R_j$. $e_j(t)$ denotes the encoding function, shown in Eq.\eqref{eq:ruleemb}. By incorporating rule-based intensity, HRTPP ensures that predictions align with medically interpretable dependencies rather than relying solely on data-driven patterns.  

\subsubsection{Numerical Feature Intensity.}  
Numerical features such as heart rate and blood pressure provide an essential context for event predictions. The numerical feature intensity component modulates event occurrence probabilities based on these continuously valued attributes. The intensity function is defined as:  
\begin{equation}  
\lambda_{\text{num}}(t) = \sum_{k=1}^K \beta_k g_k(t), 
\label{eq:num}
\end{equation}  
where $\beta_k$ is a learnable parameter representing the weight associated with predicate $X_k$. $ g_k(t) $ is the numerical feature encoder, which is shown in Eq.\eqref{eq:numemb}. For example, if a patient’s blood pressure remains high, it may increase the likelihood of a hypertension-related event. The numerical feature dynamically modulates event probabilities, providing a richer representation of patient status.  

The overall intensity function integrates all three intensity components. In order to ensure that the overall intensity function remains positive, we apply the softplus transformation. The softplus function is defined as: $\text{Softplus}(x) = \gamma \log(1 + \exp(x / \gamma)),$ where $\gamma>0$ is a learnable parameter. By incorporating Eqs.\eqref{eq:ruleemb}-\eqref{eq:num}, the final expression for the overall intensity function is as follows:

\begin{equation}
\begin{split}
\lambda(t|\mathcal{H}_t) &= \text{Softplus} \left( \lambda_{\text{base}}(t) + \lambda_{\text{rule}}(t) + \lambda_{\text{num}}(t) \right)\\
&= \text{Softplus} \left( \lambda_0 + \sum_{R_j \in \mathcal{R}} \alpha_j \sum_{t_j \in \mathcal{T}j} d{_\text{rule}}(t - t_j) + \sum_{k=1}^K \beta_k m_k \sum_{i:k_i=k} v_i d_{\text{num}}(t - t_i) \right).
\end{split}
\label{eq:overall}
\end{equation}
where $\boldsymbol{\alpha}=\{\alpha_j\}_{j=1}^{|\mathcal{R}|}$ is the set of weights associated with temporal rules, $\boldsymbol{\beta}=\{\beta_i\}_{i=1}^{K}$ is the set of weights associated with numerical features. The learnable parameters set is denoted as $\Theta=\{\boldsymbol{\alpha},\boldsymbol{\beta},\gamma,\lambda_0\}$. This overall intensity function serves as the foundation for modeling event occurrences in HRTPP framework, enabling robust modeling in event prediction and rule mining.

\subsection{Training and Prediction}
To effectively learn the parameters of the HRTPP model and enable accurate event prediction, we design a training framework based on likelihood maximization and employ our learned intensity function for future event forecasting.

\subsubsection{Loss Function.}
The HRTPP model is trained by maximizing the likelihood of observed event sequences within the temporal point process framework. Given an event sequence $\mathcal{S} = {(t_i, k_i, v_i)}_{i=1}^{N}$ and a rule set $\mathcal{R}$, the likelihood function is formulated as $\mathcal{L}(\Theta) = \prod_{i=1}^{N} \lambda(t_i | \mathcal{H}_{t_i}) \exp \left( - \int_{0}^{t_N} \lambda(t | \mathcal{H}_t) dt \right)$, where $\Theta$ is the set of learnable model parameters. To facilitate optimization, we minimize the negative log-likelihood (NLL) as the loss function:

\begin{equation}
\mathcal{L}_{\text{NLL}}(\Theta) = - \sum_{i=1}^{N} \log \lambda(t_i | \mathcal{H}_{t_i}) + \int_{0}^{t_N} \lambda(t | \mathcal{H}_t) dt.
\end{equation}
This loss function ensures that the model learns to assign higher intensities to observed events while also maintaining proper event distributions over time.

\subsubsection{Event Prediction.}
HRTPP model can be used to predict the timing of future events based on the learned intensity function. Given a history $\mathcal{H}_t$ up to time $t$, the next event time $\hat{t}$ is sampled from the conditional density function:

\begin{equation}
p(\hat{t} | \mathcal{H}_t) = \lambda(\hat{t} | \mathcal{H}_t) \exp \left( - \int_{t}^{\hat{t}} \lambda(s | \mathcal{H}_s) ds \right).
\end{equation}

By minimizing the loss function, HRTPP effectively learns the underlying intensity function of event occurrences. By predicting the timing of the next event, HRTPP can capture the evolving dynamics of the sequence, offering valuable insights and data-driven support for medical diagnosis.

\subsection{Rule Mining and Optimization}
To efficiently extract temporal logic rules, a two-phase rule mining strategy is employed, incorporating rule candidate generation and Bayesian optimization.

\subsubsection{Rule Candidate Generation.}
Temporal rules are derived from historical event sequences to construct a structured search space. Without constraints, the combination of predicates and relations can be excessively large. For instance, any rule containing $m$ predicates interconnected by $(m-1)$ relations yields a combinatorial space of $K^m C^{m-1}$ possible configurations, with $K$ and $C$ being the total number of the predicates and relations respectively. To address this, rule length is restricted, and predicate filtering is applied. Firstly, rule length significantly impacts interpretability. Overly complex rules hinder comprehension, whereas single-predicate rules lack temporal expressiveness. Based on medical guidelines, a maximum of two or three predicates is imposed. Secondly, predicate filtering is conducted via pre-training using a rule-free TPP model as a baseline. Each predicate is evaluated individually, and those reducing model loss are retained. The resulting filtered predicates are used to generate candidate rules, ensuring that each rule contains at least one valid predicate. This process refines the search space, facilitating efficient rule selection.

\subsubsection{Rule Optimization via Bayesian Optimization.}
Following candidate generation, Bayesian optimization is utilized to refine the rule set and maximize predictive performance, avoiding the infeasibility of exhaustive searches. The objective is to identify the optimal rule subset $\mathcal{R}^*$ that maximizes the model log-likelihood:
\begin{equation}
\mathcal{R}^* = \arg\max_\mathcal{R} \log\mathcal{L}(\mathcal{R};\Theta),
\end{equation}
where $\log\mathcal{L}(\mathcal{R};\Theta)$ denotes the model likelihood given the rule set $\mathcal{R}$. To maintain efficiency and interpretability, the rule set size $|\mathcal{R}|$ is is constrained to a fixed constant.

Bayesian optimization iteratively updates the rule set in three steps: (1) sampling candidate subsets using a probabilistic acquisition function, (2) evaluating model performance, and (3) refining the posterior distribution to guide further searches. This approach efficiently identifies high-quality rules while mitigating overfitting and redundancy, ensuring an optimal balance between model accuracy and interpretability. Unlike brute-force search methods, Bayesian optimization efficiently explores the parameter space.

\section{Experiments}
\subsection{Experimental Setup}
\textbf{Datasets.} We use four real-world disease-specific datasets: AKI (Acute Kidney Injury), Stroke, Sepsis, and CAD (Coronary Artery Disease). These medical datasets are extracted from MIMIC-IV \cite{johnson2018mimic}, a large-scale medical database that contains electronic health records. Table\@ \ref{table1} presents the statistical information of these datasets. The number sign \# denotes a summation operation. As shown in the table, these four datasets increase in size from top to bottom. In the AKI dataset, disease progression to Phase III serves as the target event. The Stroke dataset focuses on patients transitioning from moderate to severe conditions. In the Sepsis dataset, the occurrence of low urine output is tracked as a target of organ failure. In the CAD dataset, patient mortality is defined as the target event. \\
\textbf{Baselines.} We adopted three state-of-the-art models with explicit rule-mining capabilities: TEmporal Logic rule LearnER (TELLER) \cite{li2021explaining}, Clock Logic Neural Network (CLNN) \cite{yan2023weighted}, and CLUSTER \cite{kuang2024unveiling}.\\
\textbf{Evaluation Metrics.} To assess the performance of the models, we conducted comprehensive experiments. Each dataset is randomly split into an 80\% training set and a 20\% test set. The evaluation metrics used to measure predictive performance include \textit{Negative Log-Likelihood (NLL)}, \textit{Root Mean Squared Error (RMSE)} and \textit{Rule Accuracy (Acc)}. Rule accuracy means the proportion of correct rules.

\begin{table}[!ht]
    \centering
    \setlength{\tabcolsep}{5.5pt} 
    \caption{Datasets statistics.}
    \label{table1}
    \begin{tabular}{l c c c c c}
    \toprule
    Dataset & \# Sequences & \# Types & \# Events & Seq. Length & Avg. Length\\ 
    \midrule 
    AKI &2327   &38&58k&[3, 123]&24.93 \\ 
    Stroke &4951&48&447k&[2, 2207]&97.47 \\ 
    Sepsis & 20081&66&1375k&[2, 1406]&68.46\\ 
    CAD & 12459&56&3519k&[6, 7955]&282.45\\ 
    \bottomrule
    \end{tabular}
\end{table}

\subsection{Performance Analysis}
In this section, we compare the performance of model fitting, event occurrence time prediction, and rule accuracy using three evaluation metrics. The results shown in Table\@ \ref{mainperformance} demonstrate that HRTPP outperforms baseline models. The lowest NLL indicates that HRTPP effectively captures and models uncertainty. HRTPP also achieves the lowest RMSE, suggesting precise event time predictions and enhancing reliability in medical applications. High rule accuracy of HRTPP demonstrates the extraction of medically relevant rules, improving interpretability and supporting transparent decision-making. Among the two most accurate models, HRTPP is computationally more efficient, consistently completing training in under 16 hours on all datasets, whereas TELLER requires over 48 hours.

\begin{table}[!ht]
\centering
\caption{Comparisons of main results.}
\label{mainperformance}
\renewcommand{\arraystretch}{1.3}
\scriptsize
\begin{tabular}{l|rrr|rrr|rrr|rrr}
\hline
 & \multicolumn{3}{c|}{AKI}  & \multicolumn{3}{c|}{Stroke}  & \multicolumn{3}{c|}{Sepsis} & \multicolumn{3}{c}{CAD} \\
&NLL$\downarrow$ & RMSE$\downarrow$ & Acc$\uparrow$ & NLL$\downarrow$ & RMSE$\downarrow$ & Acc$\uparrow$ & NLL$\downarrow$ & RMSE$\downarrow$ & Acc$\uparrow$ &NLL$\downarrow$ & RMSE$\downarrow$ & Acc$\uparrow$  \\
\hline
TELLER & 12.9 & 8.1 & 80\% & 21.8 & 8.4 & \textbf{100\%} & 23.4 & 4.4 & 75\% &203.8& 5.3 & 80\% \\
CLNN & 277.5 & 10.4 & 83\% & 178.9 &13.2& 70\% & 207.2 & 5.9 & 67\% & 177.1 & 4.2 & 71\% \\
CLUSTER & 79.0 & 10.4& 60\% & 36.3 & 13.2 & 65\% & 37.5 & 8.8 & 65\% & 98.9 & 10.4 & 55\% \\
HRTPP & \textbf{7.4} & \textbf{5.4} & \textbf{100\%} & \textbf{15.1} & \textbf{4.3} & \textbf{100\%} & \textbf{12.7} & \textbf{2.7} & \textbf{80\%} & \textbf{5.8} & \textbf{3.7} & \textbf{100\%} \\
\hline
\end{tabular}
\end{table}

\subsection{Rule Analysis}
In this section, we analyze the quality of the rules from the perspectives of rule expression, correctness, reliability, and distribution, thereby verifying the interpretability of the model. We take the CAD dataset as an example. Table\@ \ref{tab:all_rules} shows top 10 rules of four models. The full results are shown in the supplementary material.

\begin{table}[!ht]
\centering
\renewcommand{\arraystretch}{1.3} 
\caption{Rules on the CAD dataset. The predicates in green boxes are direct indicators, and those in blue boxes are indirect indicators.}
\label{tab:all_rules}
\scriptsize 
\begin{tabular}{c | p{8.4cm} | c | c}
\toprule
\noalign{\vskip -3pt}
\textbf{Model} & \textbf{Rule} & \textbf{Weight} & \textbf{Correctness} \\
\noalign{\vskip -3pt}
\midrule
\noalign{\vskip -3pt}
\multirow{5}{*}{TELLER} & Respiratory Rate High \textrightarrow{} Dead & 0.2859 & Correct \\ 
 & \colorbox{green!40}{Arterial Blood Pressure diastolic Low} \textrightarrow{} Dead & 0.2202 & Correct \\ 
 & \colorbox{green!40}{Heart Rate High} \textrightarrow{} Dead & 0.1614 & Correct \\ 
 & \colorbox{green!40}{Arterial Blood Pressure mean Low} \textrightarrow{} Dead & 0.1400 & Correct \\ 
 & O2 saturation pulseoxymetry Low \textrightarrow{} Dead & 0.1283 & Incorrect \\ \hline
\multirow{7}{*}{CLNN} & \colorbox{blue!40}{c\_Glucose High} - c\_Dead \textgreater{} -0.25 & 1.03 & Correct \\ 
 & c\_BUN High - c\_Dead \textgreater{} -0.63 & 1.01 & Correct \\ 
 & c\_INR Low - c\_BUN High \textgreater{} 0.01 & 0.99 & Incorrect \\ 
 & \colorbox{blue!40}{c\_Glucose High} - c\_AST Low \textgreater{} -0.08 & 0.99 & Incorrect \\ 
 & \colorbox{blue!40}{c\_Albumin High} - \colorbox{blue!40}{c\_Glucose High} \textgreater{} 0.02 & 0.97 & Correct \\ 
 & c\_BUN High - \colorbox{green!40}{c\_C-Reactive-Protein High} \textgreater{} -0.14 & 0.96 & Correct \\ 
 & c\_Brain Natiuretic Peptide High - c\_BUN High \textgreater{} 0.41 & 0.94 & Correct \\ \hline
\multirow{10}{*}{CLUSTER} & BUN Low equal Lactic Acid Low & -& Incorrect \\ 
  & Hemoglobin High equal Hematocrit High&- & Incorrect \\ 
  & Brain Natiuretic Peptide High after ALT Low& -& Correct \\ 
  & \colorbox{green!40}{Heart Rate High} before AST Low& -& Correct \\ 
  & Calcium ionized High equal Total Bilirubin High&- & Incorrect \\ 
  & CK-MB fraction High equal Calcium ionized Low&- & Correct \\ 
  & Calcium ionized Low equal \colorbox{green!40}{Cholesterol High}&- & Correct \\ 
  & Temperature Low equal Hemoglobin High& -& Correct \\ 
  & Hematocrit Low equal Total Bilirubin Low&- & Incorrect \\ 
 & Glucose Low before Anion gap Low&- & Incorrect \\ \hline
\multirow{12}{*}{HRTPP} & \colorbox{green!40}{Heart Rate Low} equal \colorbox{green!40}{Arterial Blood Pressure diastolic High} \textrightarrow{} Dead & 2.4073 & Correct \\ 
 & Anion gap High before \colorbox{green!40}{Heart Rate Low} \textrightarrow{} Dead & 1.7742 & Correct \\ 
 & \colorbox{green!40}{Heart Rate Low} equal \colorbox{blue!40}{Potassium High} \textrightarrow{} Dead & 1.6257 & Correct \\ 
 & \colorbox{blue!40}{Potassium Low} before O2 saturation pulseoxymetry Low \textrightarrow{} Dead & 1.3959 & Correct \\ 
 & Respiratory Rate Low before \colorbox{blue!40}{Potassium High} \textrightarrow{} Dead & 1.1533 & Correct \\ 
 & \raggedright\colorbox{green!40}{Arterial Blood Pressure mean Low} before \colorbox{green!40}{Arterial Blood} \colorbox{green!40}{Pressure diastolic Low} \textrightarrow{} Dead & 1.1312 & Correct \\ 
 & Anion gap Low equal O2 saturation pulseoxymetry Low \textrightarrow{} Dead & 1.1303 & Correct \\ 
 & \colorbox{blue!40}{Glucose Low} equal \colorbox{green!40}{Arterial Blood Pressure systolic Low} \textrightarrow{} Dead & 0.5603 & Correct \\ 
 & \colorbox{green!40}{Heart Rate Low} equal INR High \textrightarrow{} Dead & 0.4213 & Correct \\ 
 & Temperature High and Respiratory Rate High \textrightarrow{} Dead & 0.3194 & Correct \\ 
\noalign{\vskip -2pt}
\bottomrule
\end{tabular}
\end{table}
\normalsize
\justifying

\subsubsection{Rule Expression Analysis.} In the CAD dataset, we set mortality as the target variable and compare HRTPP with baseline models. The correctness of the rules was determined by referencing medical guidelines jointly issued by authoritative medical bodies \cite{virani20232023}. Table\@ \ref{tab:all_rules} shows the rules, weights, and correctness. The weights of HRTPP are $\alpha_j$ in Eq.\eqref{eq:overall}, demonstrating the importance of rules. Different rules are connected by logical disjunction relationships. From the experimental results, TELLER and CLNN rely on simple variables and discover few rules, missing complex physiological interactions. CLUSTER offers more flexible rules but is highly sensitive to noise, making it difficult to extract correct medical indicators. In contrast, HRTPP uncovers clinically meaningful rules, such as "if a high anion gap occurs before low heart rate, then death occurs", highlighting the link between metabolic disorders and circulatory failure. These rules enhance accuracy and align with medical logic, aiding clinical decision-making.

\begin{figure}[!ht]
    \centering
    \includegraphics[width=.9\linewidth]{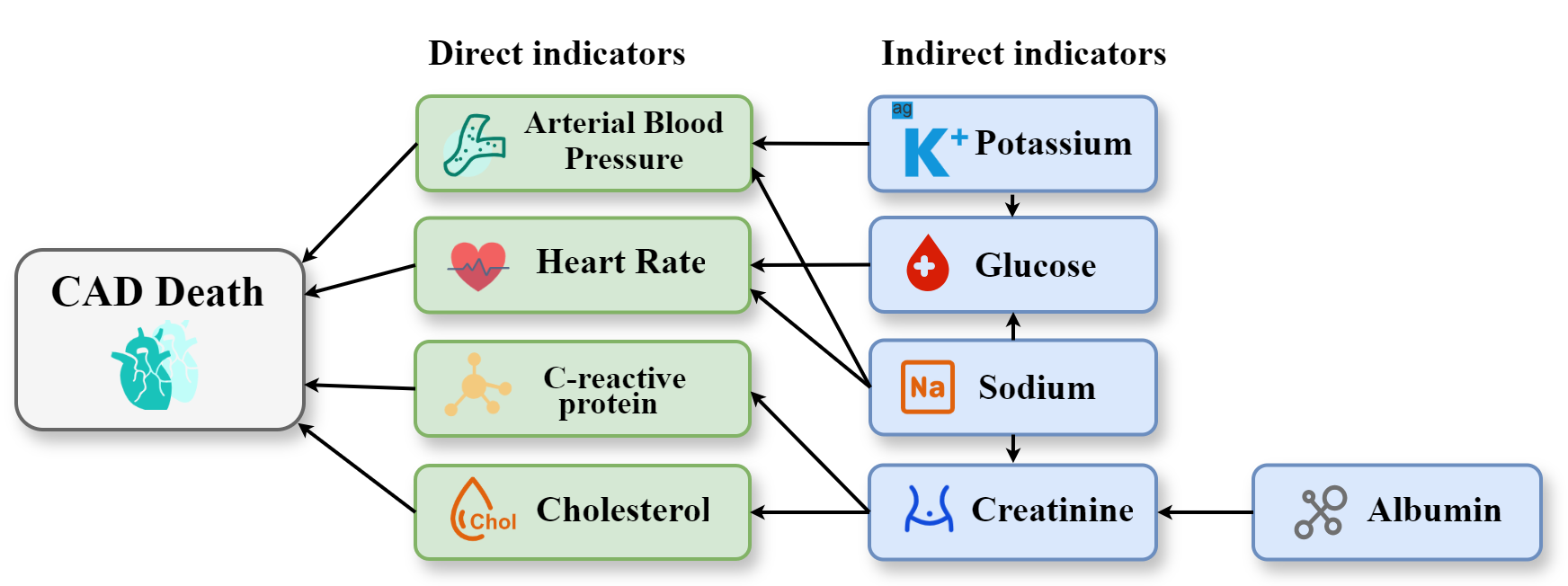}
    \caption{Hierarchical relationships of CAD indicators. The indicators in green boxes are direct indicators, and those in blue boxes are indirect indicators. Arrows denote dependencies between indicators in a cause-effect manner.}
    \label{relation}
\end{figure}
\justifying

\subsubsection{Rule Distribution Analysis.} Understanding mortality risk indicators in CAD patients is crucial for better management and prediction. To analyze these factors, we created a graph network based on medical guidelines \cite{virani20232023}, visually mapping their hierarchical relationships in Fig.\@ \ref{relation}. As shown in the Fig.\@ \ref{relation}, arterial blood pressure, heart rate, C-reactive protein and cholesterol directly effect mortality, while other indicators influence mortality through them. Hence, we classify the rules into direct or indirect rules according to their relationships. Table\@ \ref{tab:all_rules} shows direct indicators in green and indirect indicators in blue. HRTPP identifies the most direct rules, showing strong ability in capturing key CAD mortality factors. HRTPP also balances direct and indirect rules well, focusing on major factors without omitting minor details, thus enhancing its applicability in medical decision-making.

\subsection{Stability Analysis}
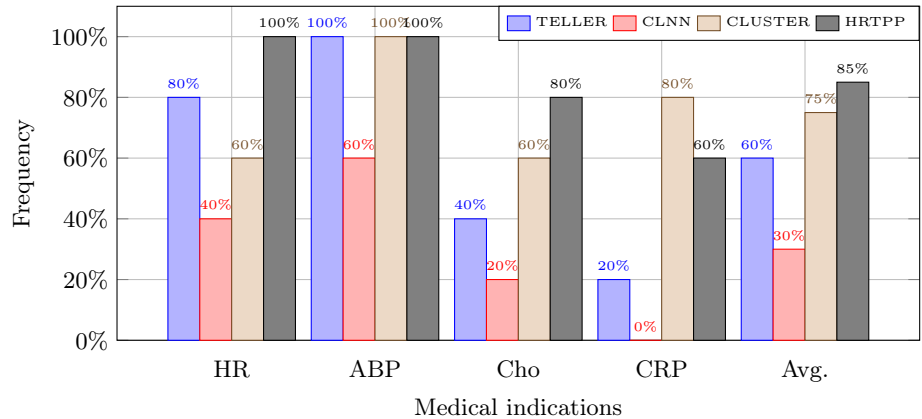
\begin{figure}[!ht]
\centering
\begin{tikzpicture}

\begin{axis}[
    ybar=0pt, 
    symbolic x coords={HR, ABP, Cho, CRP, Avg.}, 
    xtick=data, 
    xtick style={draw=none},
    grid=none,
    xlabel={Medical indications}, 
    ylabel={Frequency}, 
    ymin=0, ymax=1.1, 
    ytick={0, 0.2, 0.4, 0.6, 0.8, 1},
    enlarge x limits=0.2, 
    grid=major, 
    legend style={
        at={(1,0.949)}, 
        anchor=east, 
        legend columns=4, 
        font=\tiny 
    },
    xticklabel style={rotate=0, font=\small},
    yticklabel={\pgfmathparse{\tick*100}\pgfmathprintnumber{\pgfmathresult}\%},
    nodes near coords, 
    point meta=explicit symbolic, 
    visualization depends on={y \as \rawy}, 
    nodes near coords={
        \pgfmathparse{\rawy*100}
        \pgfmathprintnumber[fixed, precision=0]{\pgfmathresult}\%
    },
    bar width=12pt, 
    width=1\textwidth, 
    height=6cm, 
    legend image code/.code={
      \draw[#1] (0cm,-0.3em) rectangle (0.3cm,0.3em);
    },
    every node near coord/.append style={font=\tiny}
]

\addplot coordinates {(HR,0.8) (ABP,1) (Cho,0.4) (CRP,0.2)(Avg.,0.6)};
\addplot coordinates {(HR,0.4) (ABP,0.6) (Cho,0.2) (CRP,0)(Avg.,0.3)};
\addplot coordinates {(HR,0.6) (ABP,1) (Cho,0.6) (CRP,0.8)(Avg.,0.75)};
\addplot coordinates {(HR,1) (ABP,1) (Cho,0.8) (CRP,0.6)(Avg.,0.85)};

\addlegendentry{TELLER}
\addlegendentry{CLNN}
\addlegendentry{CLUSTER}
\addlegendentry{HRTPP}

\end{axis}
\end{tikzpicture}
\caption{Frequency of direct rules in five runs.}\label{stability}
\end{figure}

Model stability is crucial in medical decision support, ensuring consistency across conditions. To evaluate model stability, we recorded the frequency of four direct CAD indicators in five runs, using a five-fold cross-validation. Fig.\@ \ref{stability} presents the results, where HR, ABP, Cho, and CRP denote Heart Rate, Arterial Blood Pressure, Cholesterol, and C-reactive Protein, respectively. As shown in the figure, HRTPP has the highest frequency of direct rules, which shows the best stability and consistency. Clinically, its stability ensures interpretable results by minimizing fluctuations caused by minor data changes, thereby enhancing decision-making accuracy and supporting personalized treatment.

\justifying

\subsection{Case Study}
\begin{figure}[!ht]
    \centering
    \includegraphics[width=\linewidth, height=\textheight, keepaspectratio]{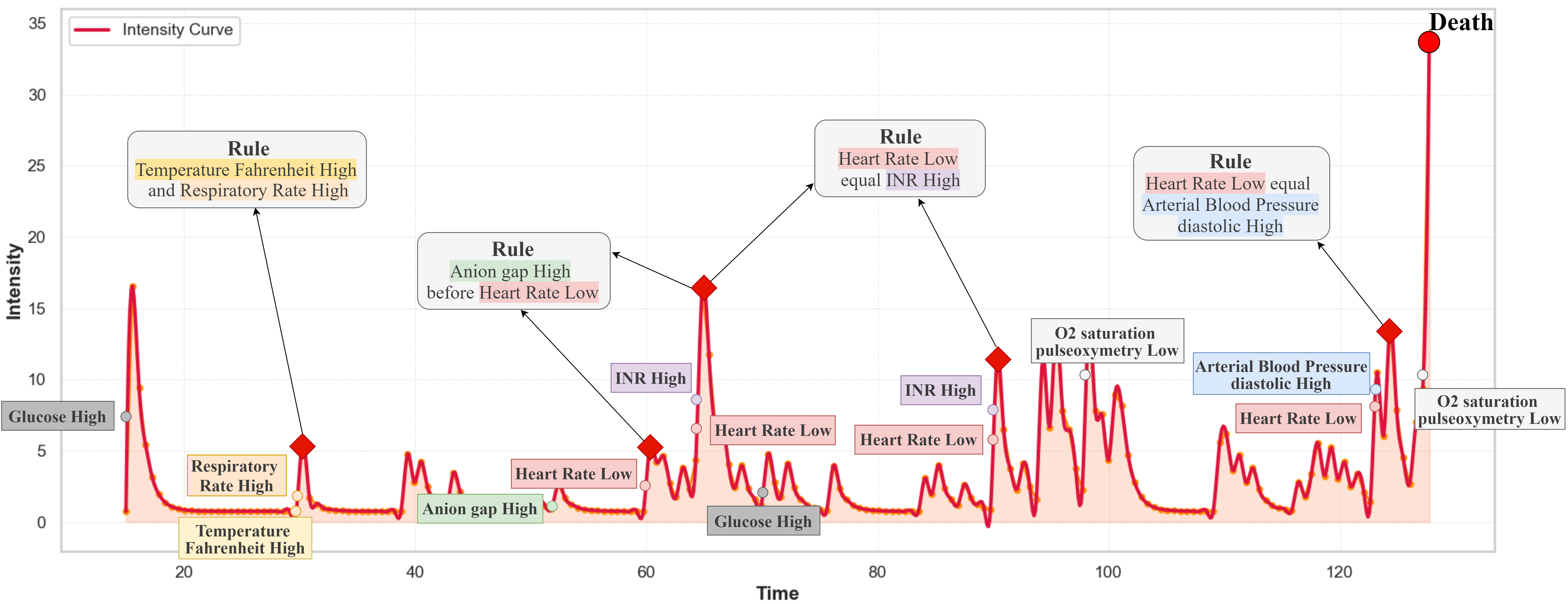}
    \caption{Overall intensity over time for a CAD patient. The red curve represents the intensity of the target variable, reflecting the probability of death. Key events affecting the mortality are annotated below the curve, while the timing of logic rules is marked above. The entire figure illustrates the patient's full clinical progression in the ICU.}
    \label{fig:enter-label}
\end{figure}

In this section, we show the mechanism by which rules and events affect the target. When a rule is triggered, its strength increases instantaneously, reflecting the potential risk associated with the condition. Fig.\@ \ref{fig:enter-label} shows the intensity curve of a CAD patient. In this figure, abnormalities in physiological indicators triggered key rules over time. At 17.19h, high blood glucose appeared, followed by infection-related signs, increasing rule strength. At 60.26h, metabolic imbalance was triggered by an elevated anion gap and low heart rate. At 92.43h, high INR (International Normalized Ratio) and low heart rate indicated coagulation abnormalities. Finally, at 123.58h, decreased heart rate and increased blood pressure triggered the end-event rule, leading to a sharp rise in intensity and death.

This mechanism enables real-time monitoring and early warning of high-risk conditions while providing interpretable analysis of disease progression to support clinical decisions. By applying rule-guided dynamic analysis, we identify key causal relationships in CAD progression and detect turning points in disease deterioration, allowing clinicians to intervene precisely at critical moments. These rules are not just data-driven predictions but verifiable medical reasoning, ensuring transparency and reliability in CAD treatment.

\subsection{Ablation Study}
We conducted an ablation study to assess the impact of Numerical Feature Augmentation (NFA) on model performance. We classify models into two categories. Category I uses only basic intensity, like probabilistic TPP models. Category II uses basic and rule-based intensity. Table\@ \ref{tab:nfa_comparison} shows the NLL of two categories. The introduction of NFA universally improve model performance across datasets, with the sole exception of AKI. The standard metrics employed in AKI evaluation underrepresent clinically significant intermediate events. This limitation is compounded by NFA focus on global rule coverage rather than domain-specific critical pattern recognition. These findings validate the effectiveness of numerical intensity in medical event modeling.

\begin{table}[!ht]
\centering
\setlength{\tabcolsep}{5.5pt}
\caption{Ablation study of NFA for NLL performance.}
\label{tab:nfa_comparison}
\begin{tabular}{lcrrrr}
\toprule
\textbf{Category} &\textbf{State} & \textbf{AKI} & \textbf{Stroke} &\textbf{Sepsis} &\textbf{CAD} \\
\midrule
\multirow{2}{*}{Category I}&w/o NFA & \textbf{7.41} & 15.26 &12.83&6.30 \\
&with NFA & \textbf{7.41} &\textbf{15.24}&\textbf{12.73}&\textbf{5.96} \\ 
\midrule
\multirow{2}{*}{Category II}&w/o NFA & \textbf{7.43} & 15.26 &12.77&6.12 \\
&with NFA & \textbf{7.43} &\textbf{15.14}&\textbf{12.72}&\textbf{5.84} \\
\bottomrule
\end{tabular}
\end{table}
\normalsize
\justifying

\section{Conclusion}
This paper proposes Hybrid-Rule Temporal Point Processes (HRTPP), a novel framework that integrates rule-based reasoning with numerical feature modeling. The proposed approach effectively incorporates domain expertise through temporal logic rules, while leveraging numerical feature encoder to capture complex event dependencies. By integrating a structured rule mining and optimization strategy, HRTPP refines the rule search process, ensuring both efficiency and robustness in modeling event sequences. Experimental results demonstrate that HRTPP achieves superior predictive accuracy and interpretability compared to baseline methods, making it a valuable tool for medical diagnosis. Future work will focus on adaptive rule learning mechanisms to improve flexibility and generalization.

\section{Acknowledgements}
This work is supported by National Natural Science Foundation of China (62406270), Shanghai Rising-Star Program (24YF2748800).


%
%
%
%
\end{document}